\documentclass[twocolumn, 10pt]{article}

\usepackage[margin=1in]{geometry}
\usepackage{amsmath, amssymb} 
\usepackage{graphicx}
\usepackage{booktabs}
\usepackage{abstract} 
\usepackage[T1]{fontenc}
\usepackage{mathptmx} 
\usepackage{authblk} 
\usepackage{hyperref}
\usepackage{graphicx}
\usepackage{tikz}
\usetikzlibrary{positioning, arrows.meta, calc, shapes.geometric}

\title{\Large \textbf{ChemFusion: A Multimodal Cross-Attention Network for Reaction Yield Prediction}}

\author[1,2]{Qiwei Han}
\author[2]{Chi Zhou}
\affil[1]{\small Department of Chemistry, Duke University, Durham, NC 27708, USA}
\affil[2]{\small Department of Computer Science, Georgia Institute of Technology, Atlanta, GA 30332, USA}
\date{}

\begin{document}
	
	\twocolumn[
	\begin{@twocolumnfalse}
		\maketitle
	\begin{abstract}
		Forecasting the outcomes of transition-metal-catalyzed reactions is notoriously complex due to the interplay of diverse physical and chemical variables. A persistent computational bottleneck has been effectively merging broad electronic descriptors with the localized, three-dimensional geometry of the reactive site. To bridge this representation gap, we present ChemFusion, a hybrid neural network that fuses conventional electronic features with explicit 3D atomic coordinates. Using a cross-attention mechanism, the model enables global electronic states to dynamically attend to specific spatial constraints within un-pooled molecular point clouds. When benchmarked against a diverse library of cross-couplings, this approach delivers exceptional predictive performance, decisively surpassing traditional single-modality frameworks. Importantly, extracting the attention matrices reveals that the architecture autonomously learns to identify and penalize restrictive steric hindrances. This provides a physically grounded interpretability, demonstrating that spatially aware networks can navigate complex reaction sterics that standard statistical models typically miss.
	\end{abstract}
	\end{@twocolumnfalse}
	]
	
	\section{Introduction}
	
	\textbf{The Challenge of Reaction Optimization.} The discovery and optimization of catalytic reactions represent a fundamental bottleneck in chemical synthesis. Navigating the vast and highly non-linear combinatorial space of substrates, catalysts, ligands, and additives has traditionally relied on empirical intuition and resource-intensive screening. While High-Throughput Experimentation (HTE) has accelerated this empirical process \cite{shevlin2017high, perera2018discovery}, the exhaustive screening of all possible reagent matrices remains physically and economically intractable. Consequently, the development of machine learning (ML) architectures capable of accurately predicting reaction yields \textit{in silico} has emerged as a transformative frontier in the molecular sciences \cite{gomez2018automatic}.
	
	\textbf{The Buchwald-Hartwig Benchmark.} Within this domain, Palladium-catalyzed Buchwald-Hartwig C--N cross-coupling serves as both a cornerstone transformation in pharmaceutical manufacturing \cite{buchwald2006cross, forhartwig2008evolution} and a premier benchmark for chemical ML models. The intricate interplay of macroscopic electronic parameters and localized steric effects in this reaction makes it notoriously difficult to model computationally. A landmark benchmark was established by the Doyle laboratory, which utilized high-throughput data from nearly 4,000 Buchwald-Hartwig reactions to train classical ML algorithms \cite{ahneman2018predicting}. By representing reactions via a tabular matrix of density functional theory (DFT) computed physicochemical descriptors (e.g., HOMO/LUMO energies, atomic charges, and dipole moments), a Random Forest model achieved a strong literature baseline for out-of-sample predictions ($R^2 = 0.9189$, RMSE = 7.84\%) \cite{ahneman2018predicting}. 
	
	\textbf{Limitations of Unimodal Baselines.} However, while this purely tabular approach successfully captures the global electronic properties governing the catalytic cycle, it fundamentally relies on macroscopic descriptors that lack continuous spatial awareness. Since the release of this dataset, numerous contemporary architectures have attempted to improve predictive performance using unimodal representations, such as SMILES-based Transformers \cite{schwaller2021predicting} and topological message-passing graph neural networks (GNNs) \cite{coley2019graph}. While these models achieve strong statistical correlations, they suffer from an inherent informational bottleneck: 1D sequences, 2D graphs, and tabular arrays mathematically collapse three-dimensional space \cite{kearnes2016molecular}. This loss of geometric fidelity inherently limits their capacity to resolve the complex steric environments of transition-metal catalytic pockets, often forcing them to operate as statistical ``black boxes'' that struggle to natively encode physical organic chemistry principles, such as ligand cone angles or spatial clashes.
	
	\textbf{The Spatial Bottleneck and Destructive Pooling.} Molecular geometries dictate the steric bulk and localized non-covalent interactions critical during transition state formation. Recently, 3D Graph Neural Networks, such as the continuous-filter convolutional neural network (SchNet), have been utilized to model continuous spatial geometries directly from atomic coordinates \cite{schutt2017schnet, schutt2018schnet}. However, integrating these 3D networks into multimodal yield prediction introduces two critical bottlenecks. First, training a highly parameterized geometric deep learning model from scratch on small-scale HTE datasets inevitably leads to representational collapse and severe overfitting. Second, standard late-fusion architectures attempt to combine modalities by applying global average pooling to the 3D graph before concatenation. This pooling mechanism irreversibly collapses the single-atom spatial resolution, often introducing mathematical noise that degrades predictive performance below that of purely tabular baselines.
	
	\textbf{The ChemFusion Architecture.} To bridge the representation gap between global electronic parameters and un-pooled 3D spatial geometries, we propose a novel multimodal architecture: ChemFusion. We systematically decouple structural feature extraction from electronic processing. To prevent spatial overfitting, we utilize a pre-trained SchNet backbone as a frozen geometric feature extractor to evaluate the 3D atomic point cloud. Concurrently, a deep Multilayer Perceptron (MLP) processes the tabular DFT-derived quantum properties to establish the overarching macroscopic electronic context. 
	
	\textbf{Cross-Attentive Fusion.} Crucially, to bypass the information bottleneck introduced by global pooling, we introduce a Multi-Head Cross-Attention Fusion Layer \cite{vaswani2017attention}. Inspired by the physical reality of chemical reactivity---where macroscopic electronic properties dictate how molecules dynamically interact within their localized spatial environments---our network utilizes the tabular electronic MLP embeddings as ``Queries'' to directly attend to the un-pooled 3D structural embeddings (``Keys'' and ``Values''). This dynamic cross-attention mechanism effectively maps the global electronic context directly onto specific steric active sites without sacrificing spatial resolution.
	
	\textbf{Contributions.} In this work, we demonstrate that this spatio-electronic multimodal approach significantly advances predictive capabilities beyond traditional unimodal paradigms. By preserving 3D spatial resolution, our cross-attentive model actively identifies and penalizes severe geometric steric clashes, achieving state-of-the-art predictive performance on the Buchwald-Hartwig benchmark dataset (RMSE = 6.22\%, $R^2$ = 0.9455). Furthermore, by extracting the learned attention weights post-training, we establish a robust framework for physical interpretability. We provide mathematical evidence detailing how the network autonomously bifurcates its attention to evaluate catalytic anchors and steric hindrance, demonstrating strong alignment with foundational human chemical intuition.

	\section{Architecture and Methodology}
	\begin{figure*}[t!]
		\centering
		\resizebox{1.0\textwidth}{!}{
			\begin{tikzpicture}[
				node distance=2cm and 2.5cm,
				block/.style={rectangle, draw, rounded corners, minimum width=3.8cm, minimum height=1.6cm, align=center, font=\small\sffamily, thick},
				arrow/.style={-Latex, thick},
				spatial/.style={block, fill=blue!5, draw=blue!60!black},
				tabular/.style={block, fill=orange!5, draw=orange!60!black},
				fusion/.style={block, fill=green!5, draw=green!60!black},
				head/.style={block, fill=gray!10, draw=black},
				labelnode/.style={font=\scriptsize\sffamily\itshape, text=black!80}
				]
				
				\node[spatial] (geom) {\textbf{3D Spatial Pathway}\\(Frozen SchNet)\\[0.15cm] \textit{Un-pooled Matrix} $\boldsymbol{H}_{3D}$};
				
				\node[tabular, below=2cm of geom] (elec) {\textbf{1D Electronic Pathway}\\(Deep Tabular MLP)\\[0.15cm] \textit{Context Vector} $\boldsymbol{h}_{elec}$};
				
				\node[fusion, right=2.8cm of $(geom.east)!0.5!(elec.east)$] (attn) {\textbf{Cross-Attention Layer}\\(Multi-Head + Add \& Norm)\\[0.15cm] \textit{Fused Context} $\boldsymbol{h}_{fused}$};
				
				\node[head, right=2cm of attn] (pred) {\textbf{Prediction Head}\\(Concatenation + Dense MLP)\\[0.15cm] \textbf{Predicted Yield} ($\hat{y}$)};
				
				\draw[arrow] (geom.east) -- ++(1.9,0) node[midway, above, xshift=0.8cm, font=\small\sffamily\bfseries] {Keys ($\boldsymbol{K}$), Values ($\boldsymbol{V}$)} |- ([yshift=0.4cm]attn.west);
				
				\draw[arrow] (elec.east) -- ++(1.9,0) node[midway, below, xshift=0.3cm, font=\small\sffamily\bfseries] {Query ($\boldsymbol{Q}$)} |- ([yshift=-0.4cm]attn.west);
				
				\draw[arrow] (attn.east) -- (pred.west) 
				node[midway, above, labelnode] {$\boldsymbol{h}_{fused}$};
				
				\draw[arrow] (elec.south) -- ++(0,-0.6) -| (pred.south) 
				node[pos=0.25, below, labelnode] {Skip Connection: Original Global Context ($\boldsymbol{h}_{elec}$)};
				
			\end{tikzpicture}
		}
		\caption{The ChemFusion Architecture. The extraction pathways independently process the un-pooled 3D spatial geometries (blue) and global electronic descriptors (orange). The Cross-Attention layer explicitly queries the spatial atomic matrix using the global thermodynamic state, dynamically routing structural features to the predictive head without employing destructive average pooling.}
		\label{fig:architecture}
	\end{figure*}
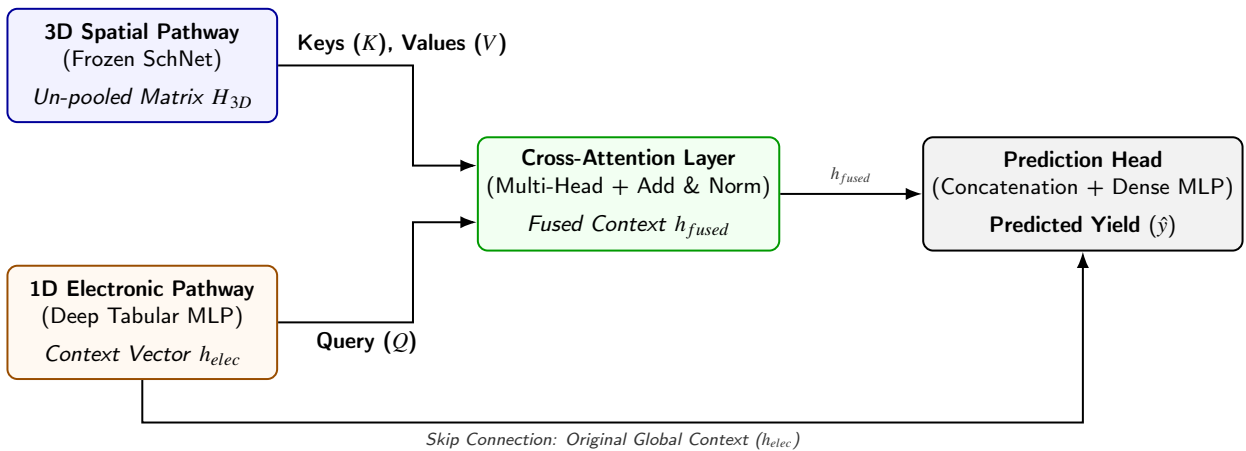
	
	\subsection{Dataset and Feature Representation}
	
	To rigorously train and evaluate our multimodal architecture, we utilized the high-throughput Buchwald-Hartwig C--N cross-coupling dataset originally published by Ahneman et al. \cite{ahneman2018predicting}. For our computational modeling, we employed the standardized, fully characterized subset comprising $N = 3,955$ experimental reaction yields. To ensure a direct and equitable comparison with prior literature baselines, this dataset and its corresponding 92 density functional theory (DFT)-derived electronic features were utilized exactly as originally reported, without supplementary mathematical modification or feature engineering.
	
	To facilitate the multimodal learning objective, each reaction is parameterized into two distinct data representations:
	\begin{itemize}
		\item \textbf{Global Electronic Descriptors:} A tabular matrix of 92 physicochemical features computed via standard DFT and cheminformatics protocols. These macroscopic descriptors encapsulate broad thermodynamic and electronic properties---such as HOMO/LUMO energy gaps, electrostatic atomic charges, and molecular dipole moments---without natively encoding continuous spatial constraints.
		\item \textbf{Explicit Spatial Geometries:} 3D atomic coordinate arrays extracted directly from the ground-state optimized geometries of the constituent substrates, ligands, bases, and additives. These arrays function as an un-pooled, mathematically continuous point cloud, providing a high-fidelity representation of the localized steric environment.
	\end{itemize}
	This bifurcated representation establishes a rigorous data foundation for the parallel extraction pathways of the ChemFusion architecture.

	\subsection{Multimodal Feature Extraction}
	
	To accurately model the highly non-linear chemical space of these cross-coupling reactions, our architecture independently processes two parallel, decoupled data streams: explicit 3D molecular geometries and macroscopic electronic descriptors, as illustrated in Figure \ref{fig:architecture}. 
	
	\textbf{Unified Spatial Complex Construction.} A fundamental challenge in modeling multi-component catalytic reactions is representing discrete molecular entities (e.g., substrates, catalysts, ligands, and additives) in a unified geometric space. Rather than processing these components independently—which precludes the network from natively learning inter-molecular physical interactions—we mathematically treat the entire reaction mixture as a single ``reaction super-complex.'' For a given reaction, the atomic numbers ($\boldsymbol{Z}$) and 3D Cartesian coordinates ($\boldsymbol{R}$) of all constituent molecules are concatenated along the sequence dimension:
	\begin{equation}
		\boldsymbol{Z}_{seq} = [\boldsymbol{Z}_{sub} \parallel \boldsymbol{Z}_{cat} \parallel \boldsymbol{Z}_{lig} \parallel \boldsymbol{Z}_{add}]
	\end{equation}
	\begin{equation}
		\boldsymbol{R}_{seq} = [\boldsymbol{R}_{sub} \parallel \boldsymbol{R}_{cat} \parallel \boldsymbol{R}_{lig} \parallel \boldsymbol{R}_{add}]
	\end{equation}
	where $\parallel$ denotes sequence concatenation. To accommodate the variable number of total atoms across disparate reactions within standard batch processing, this concatenated sequence is zero-padded to a fixed maximum length, $N_{max}$. Concurrently, a localized binary mask $\boldsymbol{m} \in \{0, 1\}^{N_{max}}$ is generated, assigning a value of $1$ to true physical atoms and $0$ to padded indices. 
	
	\textbf{Spatial Feature Extraction.} This unified geometric sequence ($\boldsymbol{Z}_{seq}$, $\boldsymbol{R}_{seq}$, and mask $\boldsymbol{m}$) is subsequently processed by a pre-trained continuous-filter convolutional neural network (SchNet) \cite{schutt2017schnet, schutt2018schnet}. To prevent severe overfitting and representational degradation on the relatively constrained high-throughput dataset, the SchNet backbone is frozen and utilized exclusively as a deterministic geometric feature extractor. Crucially, we bypass the standard global average pooling layer. By evaluating inter-atomic distance matrices via radial basis functions, the network natively outputs a continuous, un-pooled 3D spatial feature matrix, $\boldsymbol{H}_{3D} \in \mathbb{R}^{N_{max} \times d_{schnet}}$, preserving the high-dimensional spatial and electronic representation of every individual atom in the reaction mixture.
	
	\textbf{Electronic Context Extraction.} In parallel, the standardized DFT-derived electronic features \cite{ahneman2018predicting} are processed through a deep Multilayer Perceptron (MLP). This tabular processing branch incorporates Batch Normalization and Mish activation functions \cite{misra2019mish} to map the discrete physicochemical properties into a dense, macroscopic electronic latent vector, $\boldsymbol{h}_{elec} \in \mathbb{R}^{d_{elec}}$.
	
	\subsection{Cross-Attentive Fusion Mechanism}
	
	The principal architectural advancement of the ChemFusion framework lies in circumventing conventional late-fusion paradigms through the implementation of a Multi-Head Cross-Attention mechanism \cite{vaswani2017attention}. Standard multimodal approaches typically necessitate collapsing the spatial atomic matrix, $\boldsymbol{H}_{3D}$, into a singular continuous vector via global average pooling prior to concatenation. This reductive operation irreversibly degrades single-atom spatial resolution, rendering the network structurally unequipped to evaluate the localized geometric constraints that define catalytic active sites. To overcome this fundamental information bottleneck, we formulate the modality fusion not as a static concatenation, but as a dynamic, attention-driven mapping across the un-pooled 3D atomic point cloud.
	
	Specifically, the cross-attention layer is formulated using $h$ parallel attention heads. This multi-head dimensional expansion is a critical architectural choice, as it provides the mathematical capacity for feature disentanglement. Rather than forcing a single attention matrix to average all physical interactions, utilizing $h$ distinct projection spaces allows individual heads to autonomously specialize into specific chemical roles---such as anchoring to the catalytic center versus evaluating diffuse steric bulk.
	
	For mathematical clarity, the operation of a single attention head $i$ is defined as follows. The mechanism projects the incoming latent representations into three distinct matrices: the Query ($\boldsymbol{Q}_i$), the Key ($\boldsymbol{K}_i$), and the Value ($\boldsymbol{V}_i$). In our cross-modal formulation, the macroscopic electronic context vector acts as the Query, while the un-pooled spatial atomic matrix serves as the Keys and Values:
	\begin{equation}
		\boldsymbol{Q}_i = \boldsymbol{h}_{elec} \boldsymbol{W}^Q_i 
	\end{equation}
	\begin{equation}
		\boldsymbol{K}_i = \boldsymbol{H}_{3D} \boldsymbol{W}^K_i, \quad \boldsymbol{V}_i = \boldsymbol{H}_{3D} \boldsymbol{W}^V_i
	\end{equation}
	where $\boldsymbol{W}^Q_i \in \mathbb{R}^{d_{elec} \times d_{k}}$, and $\boldsymbol{W}^K_i, \boldsymbol{W}^V_i \in \mathbb{R}^{d_{schnet} \times d_{k}}$ are the learnable projection weight matrices for head $i$. 
	
	The contextualized spatial representation for the head is then computed using scaled dot-product attention:
	\begin{equation}
		\text{head}_i = \text{softmax}\left( \frac{\boldsymbol{Q}_i\boldsymbol{K}_i^T}{\sqrt{d_k}} \right) \boldsymbol{V}_i
	\end{equation}
	where $d_k$ is the dimensionality of the Query and Key vectors. Division by $\sqrt{d_k}$ acts as a temperature scaling factor to mitigate softmax saturation and prevent subsequent gradient vanishing. To ensure the network strictly evaluates physically meaningful geometries, the binary array $\boldsymbol{m}$ is applied as a key-padding mask prior to the softmax operation, forcing the attention weights of the zero-padded indices to zero. 
	
	Physically, this masked softmax operation computes the dot-product similarity between the global electronic requirements ($\boldsymbol{Q}_i$) and the individual atomic spatial environments ($\boldsymbol{K}_i$), returning normalized attention weights exclusively for the true $N$ atoms in the reaction complex. These localized weights are multiplied by the geometric values ($\boldsymbol{V}_i$) to produce a dynamically weighted, context-aware 3D representation. 
	
	The outputs of all $h$ parallel heads are then concatenated and linearly projected. Finally, this multimodal output is stabilized via a residual connection and layer normalization, concatenated with the original overarching electronic vector, and passed through a final predictive MLP to yield the calculated reaction yield ($\hat{y}$):
	\begin{equation}
		\boldsymbol{h}_{fused} = \text{LayerNorm}\Big(\text{Concat}(\text{head}_1, \dots, \text{head}_h)\boldsymbol{W}^O + \boldsymbol{Q}\Big)
	\end{equation}
	\begin{equation}
		\hat{y} = \text{MLP}([\boldsymbol{h}_{fused} \parallel \boldsymbol{h}_{elec}])
	\end{equation}
	where $\boldsymbol{W}^O$ is the final output projection matrix, and $\parallel$ denotes vector concatenation.
	
	\subsection{Physical Justification of the Attentive Inductive Bias}
	
	While the mathematical formulation of cross-attention is well-established in deep learning, its application to catalytic yield prediction warrants a rigorous physical justification. According to classical transition state theory, the activation energy barrier of a catalytic cycle is dictated simultaneously by the precise 3D steric alignment of the reactants and the stabilizing overlap of their frontier molecular orbitals \cite{forhartwig2008evolution, reid2018predictive}. 
	
	Crucially, these geometric and electronic factors are inextricably linked. The steric penalty or advantage of a localized structural feature---such as the cone angle or bite angle of a bulky phosphine ligand---cannot be evaluated independently of its overarching electronic context \cite{tolman1977steric, cavallo2016sambvca}. Its impact on catalytic success fundamentally depends on the electronic demand of the transition-metal center and the nucleophilicity of the coupling partners. Standard unimodal or late-fusion architectures often struggle to natively capture this interdependence, as they traditionally process geometric and electronic representations in decoupled mathematical spaces prior to a final linear aggregation.
	
	Our cross-attentive fusion mechanism addresses this representation gap by introducing a physically grounded inductive bias. By explicitly designating the global DFT-computed electronic parameters as the Query ($\boldsymbol{Q}$), the architecture structurally mandates that the localized 3D molecular graph (the Keys, $\boldsymbol{K}$, and Values, $\boldsymbol{V}$) is evaluated strictly through the lens of the reaction's overarching thermodynamic and kinetic requirements. 
	
	Mechanistically, this framework mathematically mirrors fundamental physical organic chemistry: the global electronic state dynamically evaluates the 3D spatial environment to identify atomic configurations that will either sterically stabilize or destabilize the activated complex. The resulting attention weights act as a learned, continuous masking function. They autonomously amplify the spatial features of the catalytic pocket when electronic conditions demand tight coordination, while gracefully attenuating remote spectator atoms (e.g., distal aliphatic chains) that do not actively participate in the rate-determining step. 
	
	The specific directionality of this cross-attention mapping is highly intentional. By defining the macroscopic electronic state as the Query, the network utilizes a global thermodynamic context to continuously evaluate a localized atomic map, smoothly condensing the $N_{max}$-atom spatial matrix into a fixed-size, contextually weighted vector. If the directionality were reversed---with localized spatial atoms independently querying global electronic constants---the model would mathematically imply that isolated spectator atoms possess the physical agency to evaluate macroscopic thermodynamic states. Such an inverted approach lacks both physical justification and computational efficiency. Consequently, the resulting fused embedding, $\boldsymbol{h}_{fused}$, functions not as a standard statistical concatenation of disparate datasets, but as an explicitly contextualized, differentiable mathematical representation of the transition state environment.
	
	\begin{table*}[t!] 
		\centering
		\caption{Model Performance and Ablation Study on Out-of-Sample Test Set}
		\label{tab:results}
		\renewcommand{\arraystretch}{1.15}
		\begin{tabular}{l l c c}
			\hline
			\textbf{Model Category} & \textbf{Architecture} & \textbf{RMSE (\%)} & $\boldsymbol{R^2}$ \\
			\hline
			Original Literature Benchmark & Random Forest \cite{ahneman2018predicting} & 7.84 & 0.9189 \\
			\hline
			Purely 3D (Spatial) & SchNet (Global Pooling) & 21.66 & 0.3376 \\
			Purely 2D (Tabular) & Deep Tabular MLP & 7.37 & 0.9232 \\
			Multimodal (Late Fusion) & Simple Concatenation & 7.92 & 0.9113 \\
			\textbf{Multimodal (ChemFusion)} & \textbf{Cross-Attention (Ours)} & \textbf{6.22} & \textbf{0.9455} \\
			\hline
		\end{tabular}
	\end{table*}
	
	\subsection{Evaluation Protocol and Metrics}
	
	The target variable for all predictive modeling was the experimentally recorded reaction yield, expressed as a continuous percentage (0--100\%). To rigorously evaluate the out-of-sample generalization capacity of the ChemFusion architecture, the dataset was randomly partitioned into a 70\% training set and a 30\% hold-out test set. This partition was executed using a fixed deterministic seed to ensure exact reproducibility and enable equitable comparisons against literature baselines. 
	
	Predictive performance was quantified using two standard continuous metrics: the Root Mean Square Error (RMSE) to measure the absolute predictive deviation from experimental observations, and the Coefficient of Determination ($R^2$) to assess the proportion of chemical variance successfully captured by the multimodal latent space.
	
	\subsection{Training Protocol and Optimization}
	
	The ChemFusion network was implemented using PyTorch and trained end-to-end. Given the continuous nature of the target variable, the network weights were optimized by minimizing a standard Mean Squared Error (MSE) loss function. 
	
	To effectively navigate the highly non-convex loss landscape characteristic of multimodal architectures, we employed the AdamW optimizer \cite{loshchilov2017decoupled}. A weight decay parameter of $1 \times 10^{-4}$ was applied to serve as a continuous regularization mechanism against tabular overfitting. Furthermore, the learning rate was dynamically governed by a Cosine Annealing schedule with Warm Restarts ($T_0 = 10$, $T_{mult} = 2$, $\eta_{min} = 1 \times 10^{-5}$) to prevent the model from converging into suboptimal local minima. Following standard conventions for attention-based representations \cite{vaswani2017attention}, the structural hyperparameter dictating the cross-attention dimensionality was fixed at $h = 8$ parallel heads. This configuration was selected to provide sufficient representational capacity for the aforementioned chemical feature disentanglement, while preemptively mitigating the inherent risk of over-parameterization on the constrained $N=3,955$ sample size.
	
	Computationally, because the highly parameterized 3D SchNet feature extractor remained entirely frozen during training, memory overhead and processing times were significantly reduced. The active parameters---comprising the Tabular MLP, the Cross-Attention mapping layer, and the final prediction MLP---were trained utilizing Automatic Mixed Precision (AMP) on a single CUDA-enabled GPU. This optimization enabled rapid convergence, with the optimal multimodal state typically achieved within 500 epochs. Model checkpoints were iteratively saved based strictly on the minimization of the out-of-sample validation loss to ensure robust generalization.
	
	\section{Results and Discussion}
	
	\subsection{Ablation Study and Baseline Comparisons}
	
	\begin{figure*}[t!]
		\centering
		\includegraphics[width=0.95\textwidth]{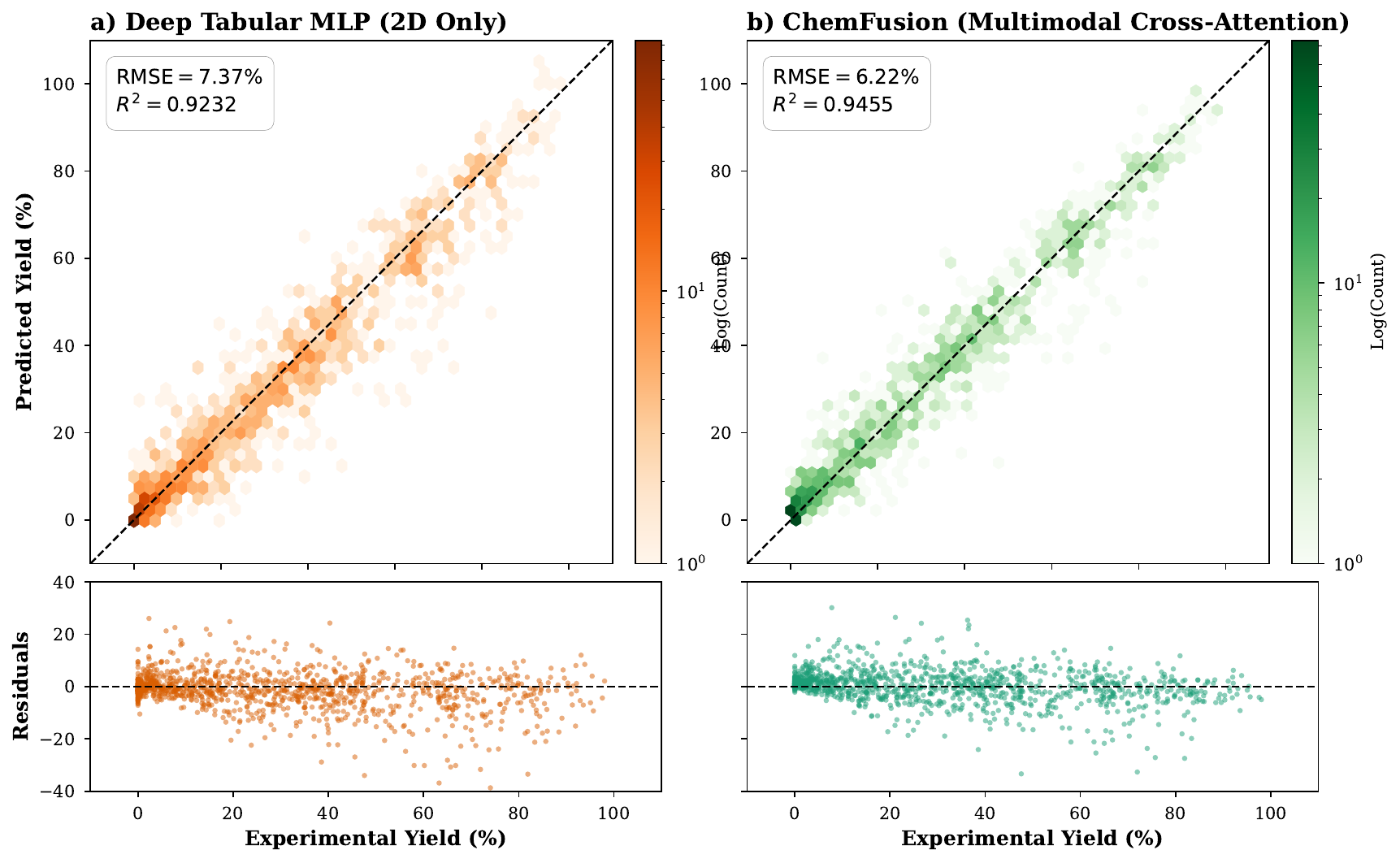}
		\caption{Visual analysis of out-of-sample predictive performance on the test set ($N \approx 1,186$). \textbf{(a)} The Deep Tabular MLP exhibits a diffuse predictive spread and a wide, heteroscedastic residual error band, struggling with structural outliers. \textbf{(b)} The ChemFusion Cross-Attention network tightly bounds predictions along the $y=x$ diagonal and features a highly compressed, zero-centered residual distribution, visually demonstrating the attenuation of systemic geometric errors.}
		\label{fig:parity}
	\end{figure*}
	
	To rigorously isolate the impact of distinct feature representations and modality fusion techniques, we performed a systematic ablation study on the out-of-sample test set. The predictive performance of all evaluated architectures is summarized in Table \ref{tab:results}.
	
	\textbf{Unimodal Baselines.} We first established performance floors utilizing strictly unimodal networks. For the spatial domain, we evaluated a purely geometric baseline utilizing the SchNet architecture. Because standard continuous regression typically requires compressing the $N \times d$ atomic graph into a single latent vector, global average pooling was applied. As anticipated, this pooling mechanism collapsed the localized single-atom spatial resolution, resulting in substantial performance degradation ($R^2 = 0.3376$, RMSE = 21.66\%). Conversely, our strictly tabular baseline—a Deep MLP trained exclusively on the 92 macroscopic electronic descriptors—achieved an $R^2$ of 0.9232 and an RMSE of 7.37\%. This unimodal electronic architecture successfully outperformed the original literature benchmark established by the Random Forest model ($R^2 = 0.9189$, RMSE = 7.84\%) \cite{ahneman2018predicting}.
	
	\textbf{Evaluating the Fusion Mechanism.} To demonstrate that our ultimate performance gains were driven specifically by the cross-attention architecture rather than the mere combination of multimodal data, we tested a standard late-fusion baseline (Simple Concatenation). In this setup, the globally pooled 3D spatial vector was directly concatenated with the macroscopic electronic latent vector before being passed to the final predictive MLP. This standard multimodal approach yielded an $R^2$ of 0.9113 and an RMSE of 7.92\%, performing slightly worse than the unimodal tabular baseline. This empirical drop demonstrates that simple concatenation can ultimately bottleneck performance; by relying on global pooling, it irreversibly collapses the spatial resolution necessary to evaluate localized steric constraints, effectively introducing mathematical noise into the latent space rather than actionable geometric context.
	
	\textbf{Cross-Attentive Superiority.} A significant architectural breakthrough occurred with the implementation of the Multi-Head Cross-Attention Fusion layer. By entirely circumventing the need for global spatial pooling and dynamically routing the macroscopic electronic query through the un-pooled 3D atomic landscape, the ChemFusion model achieved a state-of-the-art $R^2$ of 0.9455 and an RMSE of 6.22\%. 
	
	This predictive performance significantly surpasses the original literature benchmark. Furthermore, the true advantage of this multimodal framework extends beyond pure statistical accuracy into architectural robustness. While contemporary neural network approaches---such as SMILES-based Transformers \cite{schwaller2021predicting} or topological 2D GNNs \cite{coley2019graph}---have achieved remarkable success by abstracting molecules into sequences and graphs, they are inherently constrained from natively evaluating the continuous 3D spatial coordinates required to explicitly model localized steric interactions. ChemFusion bridges this representation gap by actively linking global thermodynamic properties with exact 3D atomic coordinates. By utilizing cross-attention to dynamically map macroscopic property queries onto continuous spatial structures, our model maintains high predictive fidelity even in sterically constrained regimes where traditional unimodal networks systematically struggle.
	
	\subsection{Predictive Performance and Error Distribution}
	
	To explicitly deconstruct the predictive advancements achieved by the cross-attention architecture, we analyzed the out-of-sample predictions via hexbin density parity and residual distributions (Figure \ref{fig:parity}). This multidimensional error analysis assesses whether the network's low aggregate error is structurally consistent across all reaction regimes, actively guarding against systemic biases or localized failure modes \cite{tropsha2010best}.
	
	For the strictly tabular MLP baseline (Figure \ref{fig:parity}a), the hexbin parity plot exhibits a distinctly diffuse predictive spread, particularly within the intermediate yield regime (20\%--80\%) where target variance is highest. The corresponding residual distribution displays notable heteroscedasticity, with the vertical error band frequently exceeding $\pm 30\%$. This pronounced dispersion suggests inherent limitations in relying exclusively on macroscopic electronic descriptors. Because it is structurally unequipped to natively evaluate localized steric clashes, the unimodal network appears to systematically mispredict geometrically constrained reactions, resulting in a wide footprint of structural outliers.
	
	In contrast, the ChemFusion multimodal network (Figure \ref{fig:parity}b) demonstrates high predictive fidelity. The density mapping illustrates a strict concentration of predictions tightly bounded along the ideal $y=x$ diagonal, effectively attenuating the mid-yield dispersion. Crucially, the ChemFusion residual plot exhibits a highly compressed, zero-centered homoscedastic distribution. This suggests that by dynamically mapping electronic queries against the un-pooled 3D atomic landscape, the architecture successfully resolves the specific structural outliers that historically constrained unimodal baselines. Ultimately, this visual evidence strongly indicates that the reduction in RMSE (from 7.37\% to 6.22\%) stems from a physically grounded resolution of complex reaction geometries, rather than standard statistical smoothing.
	
	\subsection{Mechanistic Interpretability via Attention Distributions}
	
	\begin{figure}[h]
		\centering
		\includegraphics[width=1.0\columnwidth]{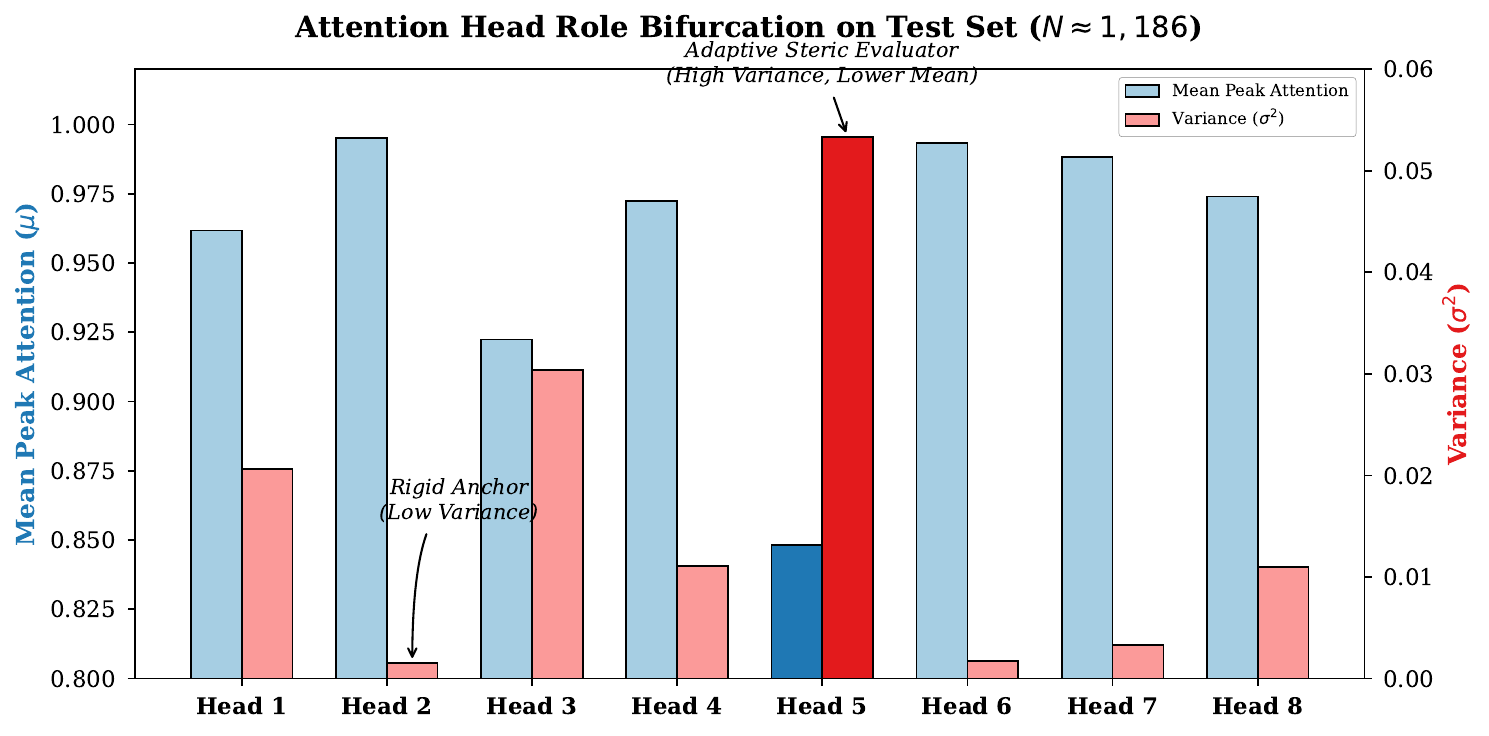}
		\caption{Statistical distribution of peak attention weights across the eight attention heads on the out-of-sample test set. The network exhibits an unsupervised bifurcation of chemical roles: Heads 2 and 6 act as rigid spatial anchors with negligible variance, while Head 5 acts as an adaptive steric evaluator, exhibiting significant spatial diffusion (high variance) to resolve complex geometric constraints.}
		\label{fig:attention_stats}
	\end{figure}
	
	A persistent challenge in applying deep learning to chemical systems is the traditionally opaque nature of highly parameterized architectures. However, the ChemFusion framework offers a pathway to explicit physical interpretability through the analysis of its learned Multi-Head Attention matrices. To investigate whether the observed performance leap was driven by chemically grounded feature extraction rather than uninterpretable statistical memorization, we analyzed the distribution of peak attention weights across the out-of-sample test set.
	
	Across all eight attention heads, the median peak attention weight converged tightly toward 1.000 (with seven heads strictly at 1.0000 and one at 0.9992). This severe concentration indicates that the network successfully avoided ``attention collapse''---a common failure mode where weights diffuse uniformly across an entire representation---and instead developed a highly localized Hard Attention strategy \cite{vaswani2017attention, ying2021transformers}. By doing so, the model dynamically routes the macroscopic electronic query through specific, heavily weighted 3D spatial coordinates.
	
	Crucially, an analysis of the mean ($\mu$) and variance ($\sigma^2$) of these peak weights across the distinct heads suggests a sophisticated, unsupervised partitioning of chemical roles. For instance, Heads 2 and 6 appear to function as ultra-rigid structural anchors ($\mu = 0.9952$ and $0.9934$; $\sigma^2 = 0.0015$ and $0.0017$, respectively). Because attention weights are bounded by a softmax probability distribution, a peak value approaching 1.0 mathematically dictates that the network has concentrated nearly all of its focus onto a single atomic node. 
	
	Remarkably, these specific heads exhibit virtually zero statistical variance across the entire, structurally diverse test set. Mathematical logic dictates that to maintain a static, zero-variance focus across thousands of unique reactions, these heads must be tracking a universally invariant spatial feature. In the mechanistic context of Buchwald-Hartwig cross-couplings, substrates, ligands, and additives are continuously varied; the solitary structurally invariant node present across all reactions is the catalytic palladium center. Consequently, we strongly hypothesize that Heads 2 and 6 persistently isolate the transition metal core, utilizing it as a static spatial reference coordinate regardless of the surrounding substrate's geometric complexity.
	
	Conversely, Heads 5 and 3 exhibited significantly lower peak means ($\mu = 0.8482$ and $0.9224$) alongside the highest test-set variances ($\sigma^2 = 0.0533$ and $0.0304$, respectively). Mathematically, a reduction in the peak maximum requires the remaining softmax probability mass to be distributed across multiple atomic nodes, indicating an adaptive spatial distribution. While these heads default to near single-atom localization for unhindered substrates, their high variance reveals a unique spatial diffusion capability. When queried with sterically hindered or complex coupling partners, Head 5 dynamically attenuates its peak focus, gracefully diffusing its attention probability across neighboring atoms to explicitly evaluate broader geometric constraints, such as bulky ligand bite angles.

	\subsection{Mechanistic Case Studies: Steric Resolution in 3D Space}
	
	\begin{figure*}[t!]
		\centering
		\includegraphics[width=0.95\textwidth]{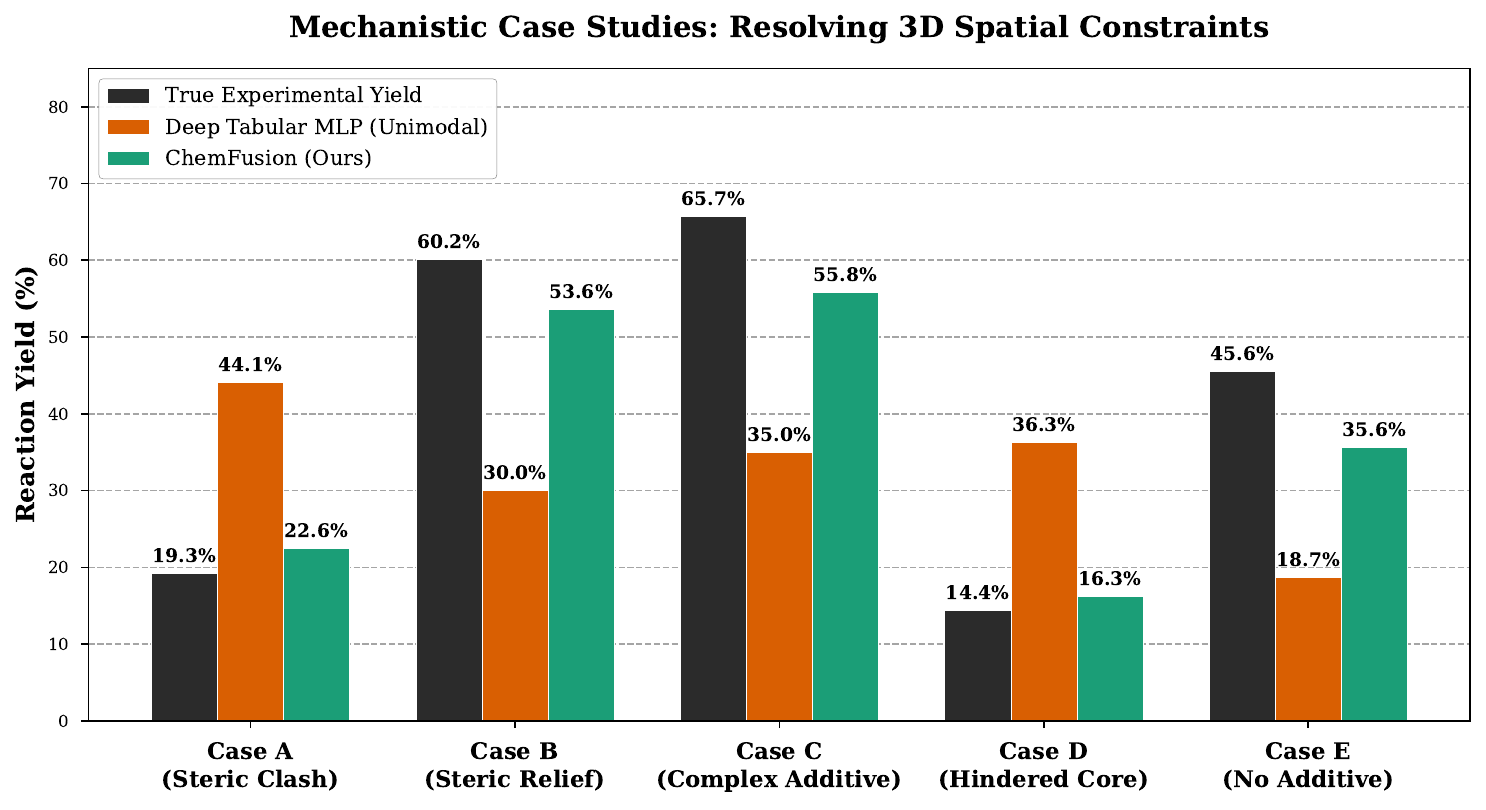} 
		\caption{Mechanistic case studies demonstrating the cross-attention network's capacity for 3D spatial resolution. Across a subset of sterically demanding reactions where the strictly tabular MLP baseline (orange) exhibits pronounced predictive dispersion, the ChemFusion architecture (green) consistently maintains high predictive fidelity. For example, in Case A (see Table \ref{tab:top_reactions}), the unimodal model is structurally unequipped to evaluate the steric clash induced by a bulky additive, resulting in a significant over-prediction. By dynamically querying the continuous 3D spatial point cloud, ChemFusion successfully accounts for these localized geometric penalties and structural reliefs, reducing absolute error margins by over 20\% on highly constrained substrates.}
		\label{fig:mechanisms}
	\end{figure*}
	
	\begin{table*}[t!]
		\centering
		\caption{Top 5 Out-of-Sample Reactions Demonstrating Cross-Attention Superiority}
		\label{tab:top_reactions}
		\renewcommand{\arraystretch}{1.2} 
		\resizebox{\textwidth}{!}{ 
			\begin{tabular}{c | l l l l | c c c}
				\hline
				\textbf{Case} & \textbf{Aryl Halide} & \textbf{Ligand} & \textbf{Base} & \textbf{Additive} & \textbf{True Yield} & \textbf{Tabular MLP} & \textbf{Cross-Attn (Ours)} \\
				\hline
				\textbf{A} & 1-bromo-4-ethylbenzene & XPhos & BTMG & 3-phenylisoxazole & 19.3\% & 44.1\% \textcolor{red}{(+24.8\%)} & \textbf{22.6\%} \textcolor{teal}{(+3.3\%)} \\
				\textbf{B} & 1-bromo-4-ethylbenzene & XPhos & MTBD & ethyl-5-methylisoxazole-3-carboxylate & 60.2\% & 30.0\% \textcolor{red}{(-30.2\%)} & \textbf{53.6\%} \textcolor{teal}{(-6.6\%)} \\
				\textbf{C} & 1-bromo-4-methoxybenzene & XPhos & BTMG & 3-methylisoxazole & 65.7\% & 35.0\% \textcolor{red}{(-30.7\%)} & \textbf{55.8\%} \textcolor{teal}{(-9.9\%)} \\
				\textbf{D} & 1-chloro-4-(CF$_3$)benzene & AdBrettPhos & MTBD & ethyl-5-methylisoxazole-3-carboxylate & 14.4\% & 36.3\% \textcolor{red}{(+21.9\%)} & \textbf{16.3\%} \textcolor{teal}{(+1.9\%)} \\
				\textbf{E} & 3-bromopyridine & XPhos & P2Et & \textit{None} & 45.6\% & 18.7\% \textcolor{red}{(-26.9\%)} & \textbf{35.6\%} \textcolor{teal}{(-10.0\%)} \\
				\hline
			\end{tabular}
		}
	\end{table*}
	
	To qualitatively validate the impact of this spatial resolution, we examined specific out-of-sample predictions where the cross-attention model significantly outperformed the unimodal baseline. Because both architectures share identical macroscopic electronic descriptors, any divergence in their predictive accuracy effectively isolates the cross-attention layer's capacity to resolve 3D geometric constraints \cite{reid2018predictive}.
	
	Among the reactions exhibiting the greatest predictive divergence (Table \ref{tab:top_reactions}), Cases A and B share an identical core substrate and ligand (1-bromo-4-ethylbenzene and XPhos) while differing exclusively in their peripheral steric environments. This structural homology provides an ideal basis for a matched-pair analysis to explicitly evaluate the model's mechanistic advantage. 
	
	In Case A, the introduction of a bulky 3-phenylisoxazole additive creates a severe geometric clash within the congested catalytic pocket. The strictly tabular MLP, being structurally unequipped to natively encode this 3D constraint, significantly over-predicts the yield at 44.1\% (True: 19.3\%). Case B represents the inverse structural scenario: the base and additive are swapped for less sterically demanding alternatives (MTBD and ethyl-5-methylisoxazole-3-carboxylate, respectively), providing spatial relief and enabling a high experimental yield of 60.2\%. Without access to these continuous spatial coordinates to evaluate the geometric relief, the tabular MLP significantly under-predicts the outcome at 30.0\%.
	
	In contrast, the ChemFusion architecture accurately captures this structural divergence. By dynamically querying the continuous 3D point cloud, the cross-attention model actively evaluates both the geometric penalty in Case A (attenuating the prediction to 22.6\%) and the structural relief in Case B (elevating the prediction to 53.6\%). This observation suggests that the multimodal fusion layer operates well beyond standard statistical aggregation. Rather, it appears to actively differentiate between sterically constrained and accessible reaction environments, mathematically adapting to localized geometric changes in regimes where macroscopic electronic descriptors alone are structurally insufficient.
	
	\subsection{Limitations and Future Directions}
	
	While the ChemFusion architecture provides a distinct advantage in resolving 3D steric constraints, we acknowledge several physical and computational limitations inherent to the current framework. 
	
	First, the model relies on static 3D atomic coordinates derived from ground-state optimizations. In experimental reality, bulky ligands such as XPhos are highly fluxional; their steric profiles exist as dynamic conformational ensembles rather than single, rigid geometries. Future iterations of this architecture could integrate molecular dynamics (MD) trajectories or multi-conformational point clouds to more accurately capture the temporal flexibility of the catalytic pocket.
	
	Second, the predictive advantages of the multimodal network inherently demand a higher computational overhead during the data generation phase. While unimodal baselines can be trained and deployed rapidly using computationally inexpensive 1D or 2D cheminformatics descriptors, ChemFusion requires the explicit geometric optimization of 3D spatial matrices for all novel substrates prior to inference. This introduces a structural preprocessing bottleneck that must be addressed before the architecture can be seamlessly deployed for massive-scale high-throughput screening (HTS).
	
	Finally, the current evaluation of this spatial resolution is confined strictly to the domain of Buchwald-Hartwig C--N cross-couplings. Within this specific mechanistic regime, the statistical variance strongly suggests that the attention heads natively anchor onto the invariant Palladium center to dynamically evaluate the surrounding steric bulk. Future studies should investigate whether this hypothesized bifurcation of attention roles generalizes to fundamentally different catalytic manifolds---such as photoredox catalysis or transition-metal-free couplings---where the definition of a structural ``anchor'' may be inherently more diffuse. Despite these scope constraints, this framework establishes a robust computational foundation for actively bridging macroscopic electronic properties with continuous three-dimensional spatial realities in chemical machine learning.
	
	\section{Conclusion}
	
	Accurately predicting transition-metal-catalyzed reaction yields remains a complex challenge. One of the most significant hurdles has been the historical difficulty of meaningfully linking global electronic properties with the 3D geometry of the catalytic pocket. To explicitly tackle this dimensional divide, we introduced ChemFusion, a multimodal deep learning architecture that bridges these distinct chemical representations via a cross-attention fusion mechanism. By systematically integrating global electronic tabular data with structurally explicit 3D point clouds, we demonstrated a significant improvement in predictive accuracy over unimodal baselines, successfully reducing out-of-sample error margins across complex Buchwald-Hartwig cross-couplings.
	
	Crucially, this performance enhancement extends beyond opaque statistical parameterization. Through the extraction and analysis of the network's multi-head attention matrices, we provided compelling mathematical evidence that the model appears to autonomously partition chemical roles. The unsupervised bifurcation of attention heads into rigid spatial anchors (tracking the catalytic center) and high-variance steric evaluators (probing ligand and additive bulk) closely mirrors established physical organic chemistry principles. Mechanistic case studies further validated this spatial sensitivity, suggesting that the cross-attention layers actively map and resolve severe geometric clashes that purely global electronic models are inherently unequipped to evaluate.
	
	Ultimately, the ChemFusion architecture establishes a robust, physically grounded framework for chemical machine learning. By moving beyond strictly "black-box" molecular representations and enabling networks to actively reconcile global electronic properties with localized spatial constraints, this architecture not only yields highly accurate predictions but also extracts interpretable insights that align with foundational chemical principles. As the field accelerates toward AI-driven catalyst design and automated synthesis, multimodal architectures that natively encode 3D spatial realities will be essential for discovering the next generation of synthetically permissible reactions.
	
	\bibliographystyle{unsrt} 
	\bibliography{references}
	
\end{document}